\documentclass{article}

\PassOptionsToPackage{authoryear,round}{natbib}
 \usepackage[preprint]{neurips_2025}

\usepackage[utf8]{inputenc} % allow utf-8 input
\usepackage[T1]{fontenc}    % use 8-bit T1 fonts
\usepackage{hyperref}       % hyperlinks
\usepackage{url}            % simple URL typesetting
\usepackage{booktabs}       % professional-quality tables
\usepackage{amsfonts}       % blackboard math symbols
\usepackage{nicefrac}       % compact symbols for 1/2, etc.
\usepackage{microtype}      % microtypography
\usepackage{xcolor}         % colors
\usepackage{graphicx}
\usepackage{svg}
\usepackage{colortbl}
\usepackage{enumitem}
\usepackage{subcaption}
\usepackage{amsthm}         % theorem environments

\theoremstyle{remark}

\usepackage{booktabs}
\usepackage{siunitx}
\usepackage{wrapfig}

\hypersetup{
    colorlinks=true,   % <-- 这是关键！设置为 true, 文本变色, 边框消失
    citecolor=neuripsaccentblue,    % 与标题颜色保持一致
    linkcolor=neuripsaccentblue,    % 与标题颜色保持一致
    urlcolor=neuripsaccentblue      % 与标题颜色保持一致
}

\definecolor{promptred}{HTML}{C70045}
\definecolor{promptbg}{HTML}{F4F4FF}

\newtcolorbox{promptbox}[1]{
  enhanced,
  breakable,
  colback=promptbg,
  colframe=promptred,
  coltitle=white,
  colbacktitle=promptred,
  title={#1},
  fonttitle=\bfseries,
  boxrule=1pt,
  arc=2mm,
  left=8pt,
  right=8pt,
  top=8pt,
  bottom=8pt,
  titlerule=0pt
}
\hypersetup{
  colorlinks=true,
  linkcolor=blue,
  citecolor=blue,
  urlcolor=blue
}

\title{ABSeeker: Training Long-Horizon Search Agents\\via Answer-Backtracked Credit Assignment}

\author{%
  \textbf{Yijun Lu\textsuperscript{1,*}}, \textbf{Rui Ye\textsuperscript{1,*,\textdagger}}, \textbf{Jiajun Wang\textsuperscript{1}}, \textbf{Yuwen Du\textsuperscript{1}}, \textbf{Tian Jin\textsuperscript{1}}, \textbf{Songhua Liu\textsuperscript{1,\textdagger}}, \textbf{Siheng Chen\textsuperscript{1,\textdagger}} \\
  \textsuperscript{1}Shanghai Jiao Tong University, \textsuperscript{*}Equal Core Contributions\\
  \textsuperscript{\textdagger}Corresponding Authors: \{yr991129, liusonghua, sihengc\}@sjtu.edu.cn
}

\begin{document}

\maketitle

\begin{abstract}
Long-horizon search agents must make multiple sequential actions (steps) to search, retrieve, verify, and integrate evidence to reach a final answer. 
However, existing methods for training these agents typically treat all steps within a trajectory uniformly during both supervised fine-tuning (SFT) and reinforcement learning (RL), failing to distinguish useful actions from erroneous or redundant ones.
In this paper, we propose \textbf{Answer-Backtracked Credit Assignment (ABC)}, a fine-grained credit assignment framework for training long-horizon search agents by converting sparse trajectory-level outcomes into dense step-level supervision that rewards useful actions (even in failed trajectories) while suppressing erroneous or redundant actions.
Specifically, given a potentially obscure query and its corresponding ground-truth answer, ABC first performs \textit{Answer-Backtracked Clue Recovery}, which traces back from the answer to recover intermediate clues required to solve the question. It then applies \textit{Clue-Anchored Step Scoring} to evaluate each search step against these clues, converting sparse binary outcome supervision into dense step-level rewards. Based on these rewards, we develop \textit{ABC-SFT}, which reweights the loss of each turn, and \textit{ABC-GRPO}, which uses the step-level scores as rewards in GRPO. Building on this framework, we train \textbf{ABSeeker} based on Qwen3.5-4B with only 8.5k examples. ABSeeker achieves 37.3\% on BrowseComp and 39.1\% on BrowseComp-ZH. With context management, the scores further improve to 55.3\% and 52.9\%, respectively, significantly outperforming same-scale (4B) agents and even matching the performance of larger ones ($\sim$30B). These results demonstrate the effectiveness of answer-backtracked step-level credit assignment for training long-horizon search agents.

\medskip
\begin{flushleft}
    \begin{tabular}{@{}ll@{}}
      \includegraphics[width=1em]{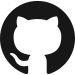} \quad \textbf{Code} & \href{https://github.com/PolarSeeker/ABSeeker}{https://github.com/PolarSeeker/ABSeeker} \\
      \includegraphics[width=1em]{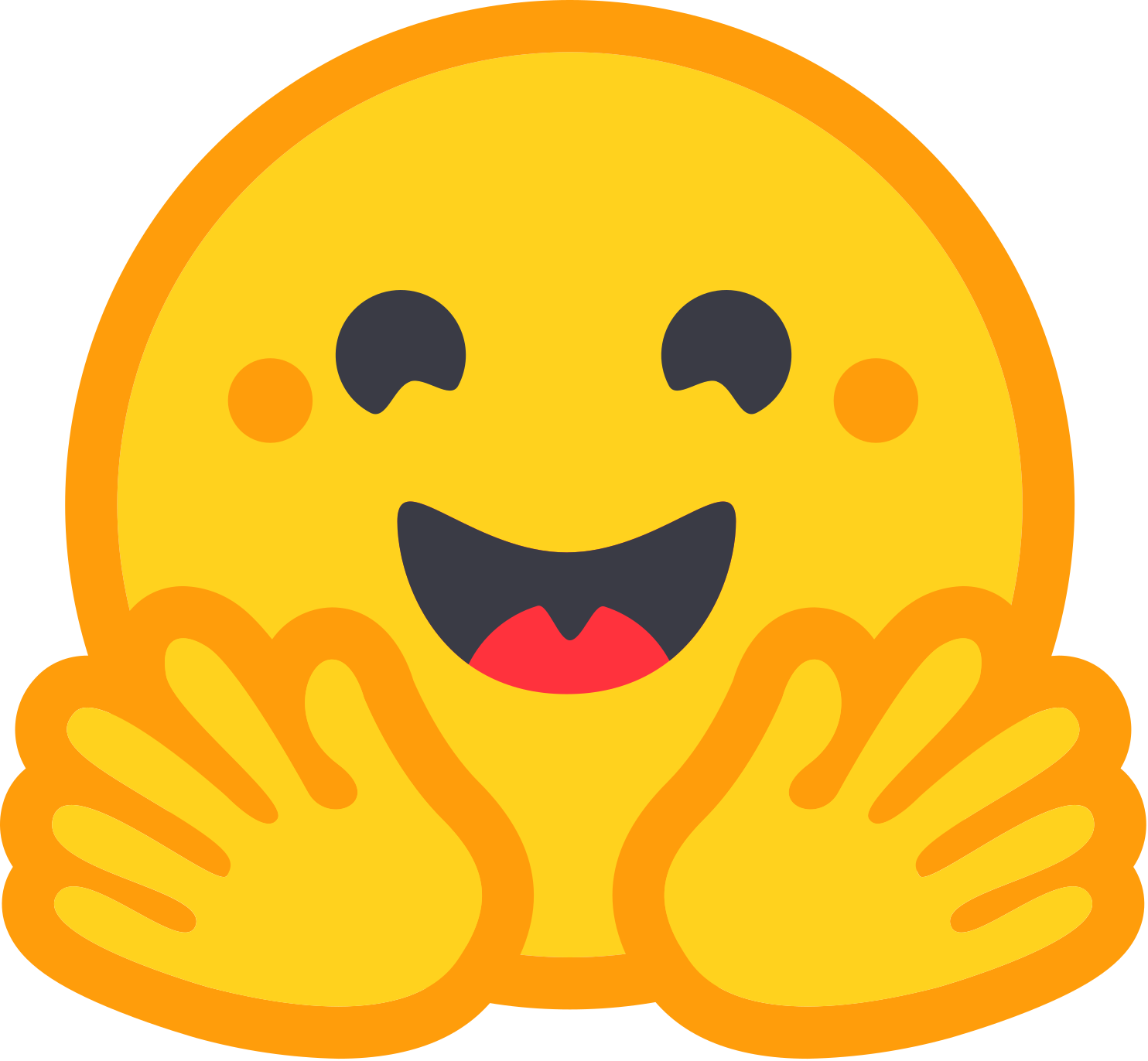} \quad \textbf{Model} & \href{https://huggingface.co/PolarSeeker/ABSeeker-4B-RL}{https://huggingface.co/PolarSeeker/ABSeeker-4B-RL} \\
    \end{tabular}
  \end{flushleft}
\end{abstract}

\begin{figure}[!h]
    \centering
    \vspace{-4mm}
    \includegraphics[width=1.0\linewidth]{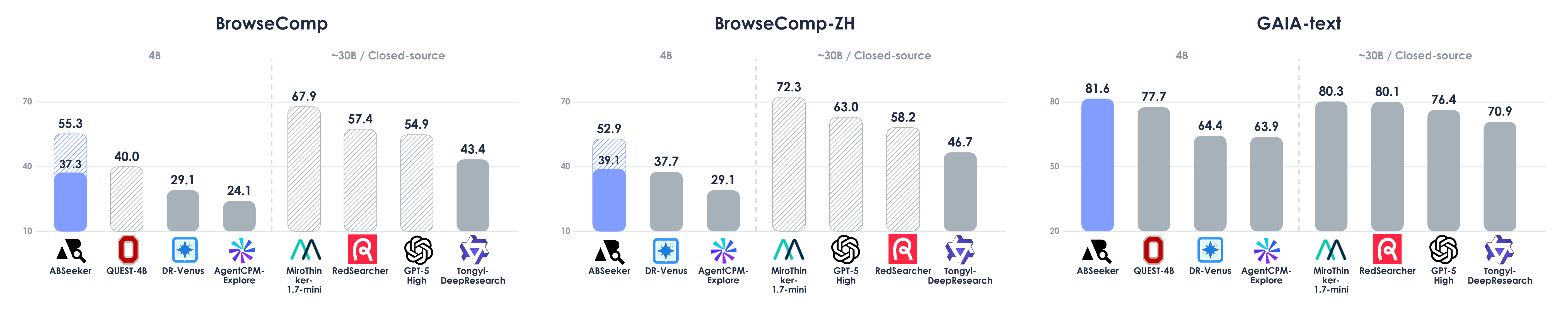}
    \vspace{-3mm}
    \caption{ABSeeker achieves the best performance among 4B models and remains competitive with several larger search agents. Striped regions indicate results with context management enabled.}
    \label{fig:teaser}
\end{figure}

\section{Introduction}
Search agents have emerged as a powerful approach for solving complex information-seeking tasks. Representative systems such as OpenAI Deep Research~\citep{openai2025deepresearch}, Tongyi DeepResearch~\citep{team2025tongyi}, and MiroThinker~\citep{team2026mirothinker} move beyond single-turn retrieval by conducting multi-step investigations, iteratively formulating queries, inspecting evidence, revising hypotheses, and adapting subsequent actions based on newly acquired information.

However, training search agents over long interaction horizons introduces a fundamental credit-assignment challenge. Existing methods typically treat all steps within a trajectory uniformly during both supervised fine-tuning (SFT) and reinforcement learning (RL), without distinguishing their individual contributions to the final outcome~\citep{jin2025search,song2025r1,chen2503research,zheng2025deepresearcher,gao2025beyond,xie2026quest}.
This uniform treatment is particularly problematic for long-horizon search, where a single trajectory often contains heterogeneous actions: even a successful trajectory may include erroneous or redundant steps, while a failed trajectory may still contain useful actions that retrieve decisive evidence or refine the solution direction. Treating all steps equally therefore fails to capture individual action quality and limits targeted optimization of critical search and reasoning behaviors.

To address this challenge, the key is to identify an effective feedback signal that can evaluate a trajectory at a fine-grained level, distinguishing the contributions of individual actions during training. 
Our core idea is motivated by a distinctive property of search tasks: once the ground-truth answer is available, the task becomes naturally backtrackable. 
Starting from the answer, one can recover the key entities, facts, relations, and constraints that should have been discovered during the search process. 
These answer-backtracked clues thus provide a principled basis for assigning differentiated credit to different parts of the trajectory.

% A distinctive property of search tasks is that, once the verified answer is available, it can be used to backtrack intermediate information that should have been discovered or correctly reasoned about during the search process. Such information can provide additional and more fine-grained supervision for training, yet it remains largely underexplored in existing search-agent methods. 

Following this idea, we propose \textbf{Answer-Backtracked Credit Assignment (ABC)}, a fine-grained credit assignment framework for training long-horizon search agents. Specifically, it consists of two stages: (1) \textit{Answer-Backtracked Clue Recovery}, which traces back from the verified answer to recover a set of intermediate clues---entities, facts, and relations that collectively characterize the evidence required to solve the question. (2) \textit{Clue-Anchored Step Scoring}, which evaluates every search step according to how it discovers, verifies, refines, or incorrectly reasons about the recovered clues, transforming sparse binary outcome supervision into dense turn-level rewards. Based on these fine-grained step rewards, we develop (1) \textit{ABC-SFT}, built upon standard SFT, which reweights the loss of each turn according to its assigned reward; and (2) \textit{ABC-GRPO}, built upon standard GRPO, which uses the step-level scores as rewards. Together, these methods provide step-level supervision that rewards useful actions even in failed trajectories, suppresses erroneous or redundant behaviors in successful ones, and enables fine-grained credit assignment through fixed answer-backtracked evaluation criteria.

Building on this framework, we train \textbf{ABSeeker} based on Qwen3.5-4B~\citep{yang2025qwen3}. ABSeeker achieves 37.3\% on BrowseComp~\citep{wei2025browsecomp}, 39.1\% on BrowseComp-ZH~\citep{zhou2025browsecomp}, 77.0\% on xbench-2505~\citep{chen2025xbench}, 46.0\% on xbench-2510~\citep{chen2025xbench}, and 81.6\% on GAIA-text~\citep{mialon2024gaia}. With context management, its performance further improves to 55.3\% and 52.9\% on BrowseComp and BrowseComp-ZH, respectively, outperforming recent same-scale 4B baselines, such as QUEST-4B~\citep{xie2026quest} and Dr.~Venus~\citep{team2026dr}, as well as larger $\sim$30B search agents such as Tongyi DeepResearch~\citep{team2025tongyi} and OpenSeeker~\citep{du2026openseeker}.

Our main contributions are summarized as follows:
\begin{itemize}[leftmargin=*, topsep=2pt, itemsep=2pt]

\item We propose \textbf{Answer-Backtracked Credit Assignment}, a fine-grained credit assignment framework that rewards useful actions in failed trajectories while suppressing erroneous actions in successful ones.

\item Based on ABC, we develop \textit{ABC-SFT}, which reweights the loss of each turn according to its step reward, and \textit{ABC-GRPO}, which incorporates step-level rewards into GRPO.

\item We first train \textbf{ABSeeker} based on Qwen3.5-4B using \textit{ABC-SFT}, and then further optimize it with \textit{ABC-GRPO}, achieving 37.3\% on BrowseComp and 39.1\% on BrowseComp-ZH.

\end{itemize}

\section{Related Work}
% ---------------------------------------------------------------
\label{sec:related}

\noindent \textbf{Search Agents.}
The ReAct paradigm~\citep{yao2022react} established the standard recipe for LLM-based web agents: interleave reasoning, tool calls, and observations to solve knowledge-intensive tasks through dynamic interaction with external environments. Recent work has scaled this framework to long-horizon search, where agents execute dozens of sequential retrieve--browse--integrate cycles to locate fine-grained information distributed across multiple sources. Representative search agents, such as OpenAI's Deep Research~\citep{openai2025deepresearch} and Tongyi DeepResearch~\citep{team2025tongyi}, have been developed for these tasks. Despite these advances, most existing agents are trained primarily with trajectory-level supervision---successful trajectories are treated as uniformly positive and failed trajectories as uniformly negative---providing limited guidance on which intermediate decisions genuinely contribute to finding the correct answer~\citep{zheng2025deepresearcher,team2026mirothinker,chu2026redsearcher,team2025tongyi,du2026openseeker}.

\noindent \textbf{Credit Assignment for Search Agents.}
Several recent methods have explored fine-grained credit assignment to address this gap. IGPO~\citep{wang2026information} assigns step-level rewards based on the increase in the model's likelihood of the ground-truth answer, but the resulting credit depends on the model's own belief estimation and fluctuates with policy updates. CSO~\citep{li2026verified} identifies critical steps by testing alternative actions and verifying whether they lead to correct outcomes. However, it assigns credit only to verified critical decisions, providing no direct supervision for the remaining steps in the trajectory. SAPO~\citep{liu2026beyond} and MindDR~\citep{team2026mind} assign step-level credit based on intermediate entities, using either their graph proximity to the answer or their coverage in the trajectory. However, such entity-level signals cannot directly determine whether each search or reasoning decision is valid. Collectively, these methods do not provide dense and reliable supervision that directly evaluates the correctness and contribution of every action throughout the trajectory.

\noindent \textbf{Our Approach.}
We propose \textbf{Answer-Backtracked Credit Assignment (ABC)}, which uses verified answers to construct dense, fine-grained process supervision. ABC first applies Answer-Backtracked Clue Recovery to recover intermediate evidence clues, and then uses Clue-Anchored Step Scoring to assign each step a scalar score based on its contribution or error. These scores reweight SFT losses in \textit{ABC-SFT} and serve as rewards for GRPO in \textit{ABC-GRPO}, from which we train \textbf{ABSeeker}.

% We take a different path by directly exploiting the verified answer associated with each training question to construct dense and fine-grained process supervision. In the first stage, \textbf{Answer-Backtracked Clue Recovery} takes a query and its verified answer and recovers a set of intermediate evidence clues, such as relevant entities, facts, and relationships, that a valid search process should establish before reaching the answer. These clues are constructed before training and explicitly define meaningful progress for each question. In the second stage, \textbf{Clue-Anchored Step Scoring} evaluates every step against the recovered clue set, determining whether it advances the evidence chain, contributes useful information, or introduces an incorrect decision, and assigns a scalar score accordingly. The resulting signal is used to reweight SFT objectives and provide turn-level rewards for RL, preserving useful exploration in failed trajectories while suppressing erroneous actions in successful ones. In this way, \textbf{Answer-Backtracked Credit Assignment} converts sparse binary outcome supervision into dense step-level credit across collected trajectories, from which we train \textbf{ABSeeker}.

% ---------------------------------------------------------------
\section{Method}
% ---------------------------------------------------------------
\label{sec:method}

\begin{figure}[t]
\centering
\includegraphics[width=\textwidth]{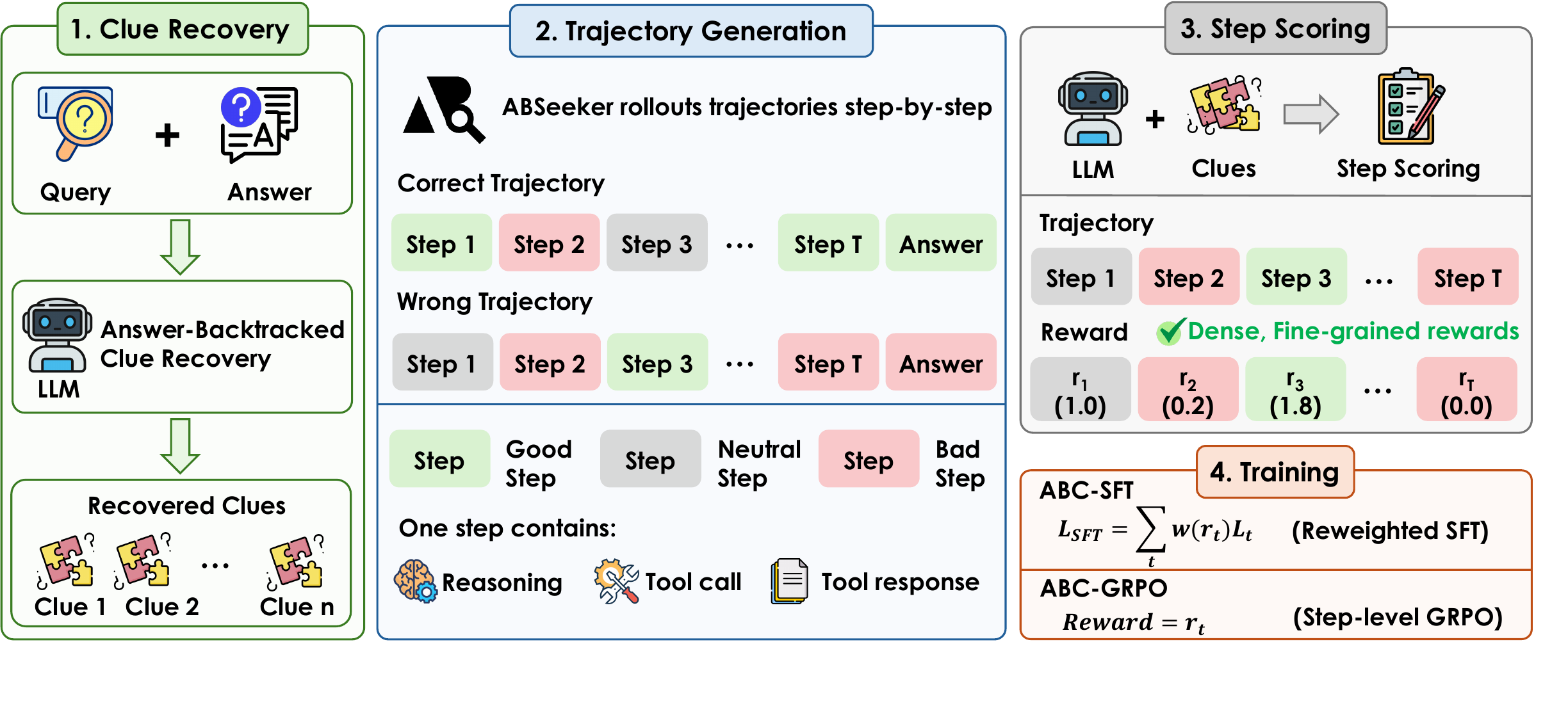}
\caption{Overview of the training pipeline, which consists of two core stages. \textbf{Stage 1: Answer-Backtracked Clue Recovery} recovers intermediate evidence clues from the query and its verified answer. \textbf{Stage 2: Clue-Anchored Step Scoring} evaluates each step against the recovered clues and produces dense, fine-grained rewards for each step, in contrast to the sparse answer-level reward.}
\label{fig:pipeline}
\end{figure}

Figure~\ref{fig:pipeline} provides an overview of the training pipeline, which consists of two core stages. Section~\ref{sec:problem} first formalizes the search trajectory and step-level credit assignment problem. Given a query and its verified answer, \textbf{Answer-Backtracked Clue Recovery} recovers a set of intermediate evidence clues that define meaningful progress toward the answer (Section~\ref{sec:clue-recovery}). Search trajectories are then rolled out, retaining both successful and failed trajectories. Next, \textbf{Clue-Anchored Step Scoring} evaluates every step against the recovered clue set and produces dense, fine-grained step scores (Section~\ref{sec:rubric}). Finally, these step-level scores directly serve as the reward signal for policy optimization (Section~\ref{sec:training}).

\subsection{Search Trajectory and Credit Assignment}
\label{sec:problem}

We consider a set of training questions, each consisting of a query $q$ and a verified answer $a^*$. A search agent interacts with a web environment over $T$ turns to produce a search trajectory:
\begin{equation}
\tau = (s_1, s_2, \ldots, s_T,\, a),
\end{equation}
where $a$ is the final answer submitted by the agent. Each step $s_t$ contains the agent's reasoning, the issued tool call, and the corresponding tool response returned by the environment.

During training, the agent typically receives a reward based solely on whether the final answer matches the ground truth:
\begin{equation}
r_{\text{ans}}(\tau) =
\begin{cases}
1, & \text{if } a = a^*,\\
0, & \text{otherwise}.
\end{cases}
\end{equation}
This trajectory-level signal is sparse and coarse, leading to two fundamental credit-assignment failures. First, an incorrect trajectory may contain several useful intermediate steps---such as correct evidence discovery, verification, or candidate filtering---yet the final reward of zero provides no positive signal for these actions. Second, a correct trajectory may contain erroneous intermediate conclusions or steps that discard useful evidence, yet the final reward of one does not distinguish these flawed actions from genuinely informative ones. In both cases, trajectory-level outcome supervision provides no per-step signal indicating which decisions advanced or impeded progress toward the verified answer.

\textbf{ABC} addresses this limitation by constructing a step reward $r_t$ for every step in each trajectory. These rewards are produced by \textbf{Clue-Anchored Step Scoring} (Section~\ref{sec:rubric}) based on the clues recovered through \textbf{Answer-Backtracked Clue Recovery} (Section~\ref{sec:clue-recovery}). The recovered clues provide fixed anchors for evaluating each rollout, enabling stable, dense, and fine-grained supervision that distinguishes useful actions from erroneous ones regardless of the final outcome.

\subsection{Answer-Backtracked Clue Recovery}
\label{sec:clue-recovery}

The verified answer $a^*$ specifies where the search should end, but provides little supervision about how the agent should reach it. To evaluate the quality of intermediate steps, we require a set of answer-backtracked anchors---concrete pieces of evidence that a valid search process should establish and that can serve as reference points for step-level evaluation. \textbf{Answer-Backtracked Clue Recovery} maps each training question $(q, a^*)$ to a set of clues:
\begin{equation}
\mathcal{C} = \{c_1, c_2, \ldots, c_K\},
\end{equation}
where each $c_k$ is a verifiable piece of intermediate evidence relevant to answering $q$, such as a specific entity, fact, attribute, or relationship connecting the query to the verified answer.

The recovery process leverages the fact that benchmarks such as BrowseComp~\citep{wei2025browsecomp} provide unique and verifiable answers, which offer a clear endpoint for evidence backtracking. Each question contains a set of constraints that, together with the verified answer, implicitly define a single valid evidence path. Given the query and its verified answer, we prompt an LLM to reconstruct the evidence chain by identifying the intermediate entities and facts that must be discovered, verified, and cross-referenced. Crucially, this backtracking is itself an active ReAct loop: the recovery model conducts web searches and visits pages through the same tool-call protocol as the forward agent, tracing evidence from the answer back toward the query and anchoring each clue in actual web content. Clues that survive this verification serve as reliable, answer-backtracked reference points for subsequent step scoring.

\begin{figure}[t]
\centering
\includegraphics[width=0.85\textwidth]{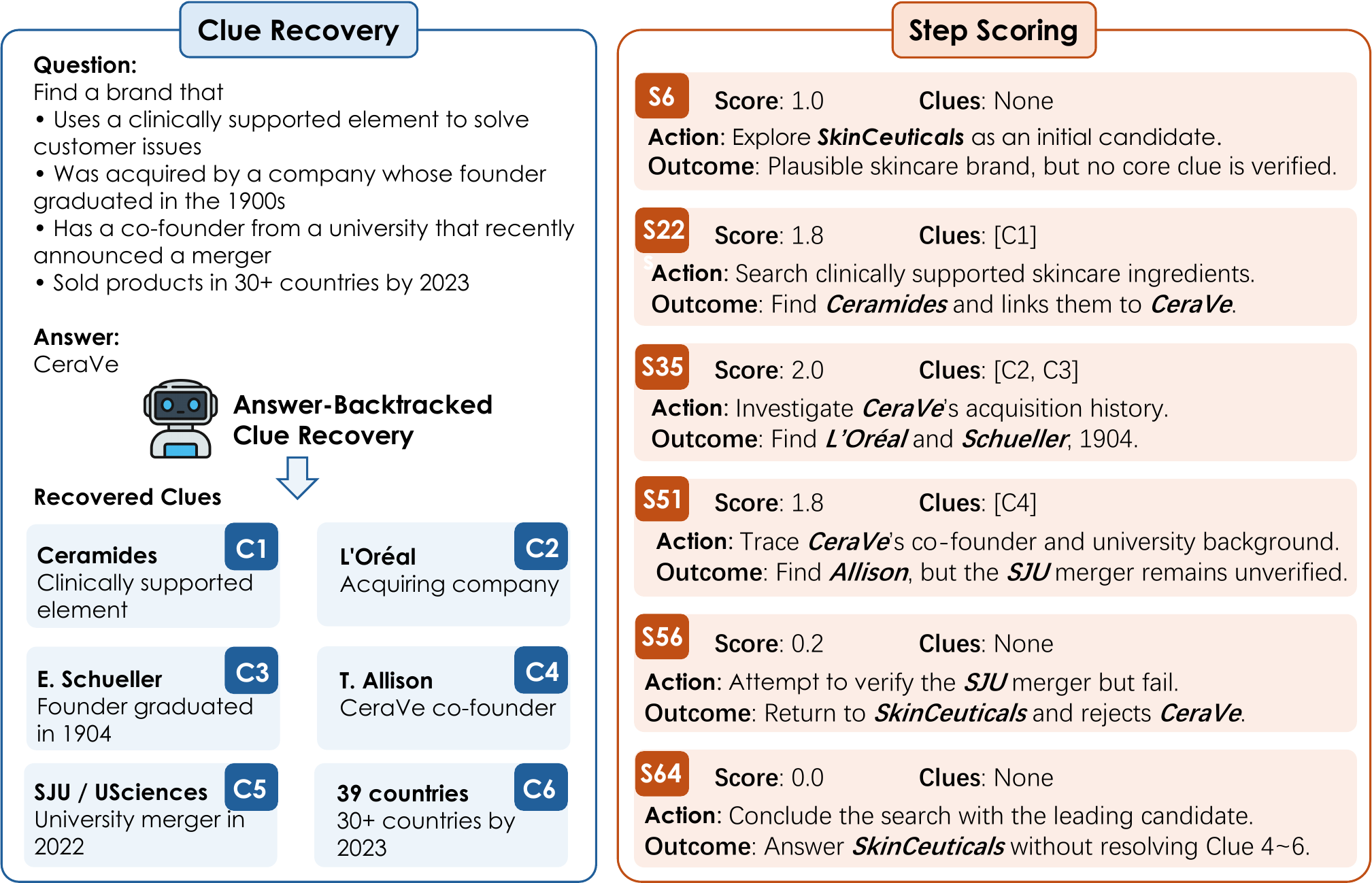}
\caption{An illustrative example of \textbf{Answer-Backtracked Clue Recovery} and \textbf{Clue-Anchored Step Scoring}. \textbf{Left:} Given a multi-constraint query and the verified answer, the recovery produces six intermediate evidence clues ($c_1$--$c_6$) that form a verified evidence chain connecting the query to the answer. \textbf{Right:} Selected steps from a sampled trajectory are evaluated against the recovered clue set, with rewards reflecting the quality and contribution of each step.}
\label{fig:example}
\end{figure}

Figure~\ref{fig:example} (left) illustrates this process using a concrete question from the training set. Given a four-constraint query and the verified answer \textit{CeraVe}, the recovery model produces six clues, including \textit{Ceramides} as the clinically supported ingredient, \textit{L'Oréal} as the acquiring company, and \textit{Eugène Schueller} as its founder who graduated in 1904. Together, these clues form a verified evidence chain connecting the query constraints to the answer. See Appendix~\ref{app:prompt-recovery} for the full recovery prompt.

\subsection{Clue-Anchored Step Scoring}
\label{sec:rubric}

Given the recovered clue set $\mathcal{C}$ for a training question, \textbf{Clue-Anchored Step Scoring} evaluates every step in each collected trajectory. For each step $s_t$, the scorer receives three inputs: (1) the current step, including its reasoning, tool call, and tool response; (2) the original query $q$; and (3) the complete clue set $\mathcal{C}$. It outputs a step reward $r_t$ together with a brief rationale explaining the applied criteria and any relevant clues. Each step starts with a base score of $1.0$, ensuring that reasonable exploration without an obvious error is not penalized. The specific scoring criteria are listed in Table~\ref{tab:rubric}.

\begin{wraptable}{r}{0.48\columnwidth}
\vspace{-8pt}
\centering
\normalsize
\caption{\textbf{Clue-Anchored Step Scoring} rubric.}
\label{tab:rubric}
\vspace{-4pt}
\setlength{\tabcolsep}{3pt}
\renewcommand{\arraystretch}{1.08}
\begin{tabular}{@{}p{0.36\columnwidth}r@{}}
\toprule
\textbf{Scored Behavior} & \textbf{$\Delta$} \\
\midrule
Discovers or verifies a correct clue
& $+0.8$ \\

Rules out an incorrect candidate
& $+0.4$ \\

Incorrectly dismisses a correct clue
& $-0.8$ \\

Submits the verified answer
& $+1.0$ \\

Submits an incorrect answer
& $-1.0$ \\
\bottomrule
\end{tabular}
\vspace{-8pt}
\end{wraptable}

A step may exhibit multiple scored behaviors, and the same behavior may occur multiple times when several clues are affected. The corresponding deltas in Table~\ref{tab:rubric} are accumulated on top of the base score and clipped to $[0,2.0]$:
\begin{equation}
r_t =
\operatorname{clip}
\left(
1.0 + \sum_{j \in \mathcal{A}_t}\Delta_j,\,
0,\,
2.0
\right),
\end{equation}
where $\mathcal{A}_t$ denotes the set of scored behavior instances detected at step $t$. A step that discovers a correct clue in a trajectory that ultimately fails still receives positive credit, whereas a step that incorrectly dismisses a correct clue in a trajectory that ultimately succeeds still receives a penalty.

Figure~\ref{fig:example} (right) illustrates the scoring process. After recovering the six clues ($c_1$--$c_6$) in Section~\ref{sec:clue-recovery}, selected steps from a sampled trajectory are evaluated against the clue set. Step~22 receives $1.8$, consisting of the base score of $1.0$ and a $+0.8$ reward for discovering \textit{Ceramides} and linking them to \textit{CeraVe}. Step~35 verifies both the \textit{L'Oréal} acquisition and \textit{Eugène Schueller}'s graduation, covering $c_2$ and $c_3$; the two positive deltas are accumulated, and the resulting score is clipped to $2.0$. Step~56 receives $0.2$ because the agent abandons the accumulated evidence supporting \textit{CeraVe} and returns to an incorrect candidate. Finally, Step~64 receives $0.0$ after the agent submits an incorrect final answer. See Appendix~\ref{app:prompt-scoring} for the full scoring prompt.

\subsection{Training with Step Rewards}
\label{sec:training}

Given the step reward $r_t$ assigned to every step in each trajectory, we train the agent in two consecutive stages: reward-weighted supervised fine-tuning (ABC-SFT) followed by step-level reinforcement learning (ABC-GRPO). Both successful and failed trajectories are retained, allowing high-quality steps to be reinforced while erroneous steps are down-weighted or penalized.

\subsubsection{Reward-Weighted Supervised Fine-Tuning (ABC-SFT)}

In the first stage, we perform SFT on all collected trajectories. For a trajectory $\tau$ of length $T$, let $x_{t,j}$ denote the $j$-th policy-generated token at step $s_t$, with environment-provided tool responses excluded from optimization. The training objective is
\begin{equation}
\mathcal{L}_{\text{SFT}}(\theta)
=
-\sum_{t=1}^{T} w(r_t)
\sum_{j}
\log p_\theta
\left(
x_{t,j}
\mid
x_{t,<j}
\right),
\label{eq:sft}
\end{equation}
where the step-level weight is computed via a sigmoid function $w(r_t) = \sigma\big(\alpha \cdot (r_t - \beta)\big)$, with $\alpha$ controlling the sharpness of the mapping and $\beta$ the neutral baseline. High-scoring steps thus contribute more strongly to the gradient, whereas low-scoring steps contribute little training signal.

\subsubsection{Step-Level Reinforcement Learning (ABC-GRPO)}

In the second stage, we further optimize the agent through online RL with step-level rewards. For each sampled rollout $i$, the reward at step $t$ is defined as
\begin{equation}
R_{i,t} = r_{i,t},
\label{eq:rl}
\end{equation}
where $r_{i,t}$ is the clue-anchored step score from Section~\ref{sec:rubric}. We normalize rewards within each rollout group to obtain $\widehat{R}_{i,t}$ and compute the discounted step-level advantage:
\begin{equation}
A_{i,t}
=
\sum_{k=t}^{T_i}
\gamma^{k-t}\,\widehat{R}_{i,k},
\label{eq:advantage}
\end{equation}
where $\gamma$ controls how future step rewards are propagated to earlier decisions. The resulting advantage $A_{i,t}$ is assigned to all policy-generated tokens at step $t$, while environment-provided tool responses are masked from optimization. We optimize the policy using the standard clipped GRPO objective~\citep{shao2024deepseekmath}, replacing its trajectory-level advantage with the step-specific advantage $A_{i,t}$.

% ---------------------------------------------------------------
\section{Experiments}
% ---------------------------------------------------------------
\label{sec:experiments}

\subsection{Experimental Setup}
\noindent \textbf{Training Setup.}
We use OpenSeeker~\citep{du2026openseeker} as the training data, collecting trajectories with both correct and incorrect final answers. The maximum number of steps per trajectory is capped at 200. We use Qwen3.5-4B~\citep{yang2025qwen3} as the backbone model.  For SFT, we train on 8.5K trajectories for 3 epochs. For RL, we sample 1000 questions, each with 8 rollouts, starting from the resulting SFT checkpoint. For both Answer-Backtracked Clue Recovery and Clue-Anchored Step Scoring, we use DeepSeek-V4-Flash~\citep{xu2026deepseek} as the backbone LLM.
Refer to Appendix~\ref{app:training} for additional details.

\noindent \textbf{Evaluations.}
We benchmark ABSeeker against the following four evaluation suites, which together span deep research, persistent web navigation, and general-purpose agent reasoning: (1) BrowseComp~\citep{wei2025browsecomp}, designed to probe long-horizon English browsing and information-seeking through complex multi-constraint queries; (2) BrowseComp-ZH~\citep{zhou2025browsecomp}, a Chinese-language counterpart that poses similarly difficult retrieval tasks on the Chinese web; (3) xbench~\citep{chen2025xbench}, which tests deep research competence---planning, reasoning, and cross-source synthesis across professional real-world scenarios; and (4) GAIA (Text-Only)~\citep{mialon2024gaia}, a suite of general assistant tasks demanding combined web browsing, tool use, and multi-hop inference. For all benchmarks the agent is allowed up to 200 tool calls. We run each evaluation three times and report the averaged score.

\noindent \textbf{Baselines.}
To validate the effectiveness of our method, we compare ABSeeker against three categories of baselines. (1) \textit{Frontier foundation models} with search capability: Gemini-3.1-Pro~\citep{google2026gemini31}, Seed-2.0-Pro~\citep{seed2026seed2}, GLM-5~\citep{zeng2026glm}, DeepSeek-V4-Pro-Max~\citep{xu2026deepseek}, and GPT-5 High~\citep{singh2025openai}. (2) \textit{Search agents at $\sim$30B}: MiroThinker-1.7-mini~\citep{team2026mirothinker}, RedSearcher~\citep{chu2026redsearcher}, DeepMiner~\citep{tang2025beyond}, Tongyi-DeepResearch~\citep{team2025tongyi}, and OpenSeeker~\citep{du2026openseeker}. (3) \textit{Search agents at 4B}: QUEST-4B~\citep{xie2026quest}, DR-Venus~\citep{team2026dr}, and AgentCPM-Explore~\citep{chen2026agentcpm}.

\subsection{Experimental Results}

\begin{table}[t]
\centering
\caption{
Performance comparison across five benchmarks.
For BrowseComp and BrowseComp-ZH, * denotes results obtained without context management.
Within each model category, the bold score denotes the best result
on each benchmark. ``--'' indicates that the result is not reported.
}
\label{tab:main-results}
\resizebox{\linewidth}{!}{%

\begin{tabular}{@{}lc*{5}{S[table-format=2.1,table-number-alignment=center]}@{}}

\toprule

\textbf{Model}
& \textbf{Param}
& \multicolumn{1}{c}{\textbf{BrowseComp}}
& \multicolumn{1}{c}{\textbf{BrowseComp-ZH}}
& \multicolumn{1}{c}{\textbf{xbench-2505}}
& \multicolumn{1}{c}{\textbf{xbench-2510}}
& \multicolumn{1}{c}{\textbf{GAIA-text}} \\

\midrule
\multicolumn{7}{c}{
    \emph{\textbf{Foundation Model with Tools}}
} \\
\midrule

Gemini-3.1-Pro
& --
& \multicolumn{1}{c}{--}
& {\bfseries 85.9}
& \multicolumn{1}{c}{--}
& 53.0
& {\bfseries 80.6} \\

Seed-2.0-Pro
& --
& 77.3
& 82.4
& \multicolumn{1}{c}{--}
& \multicolumn{1}{c}{--}
& 78.6 \\

GLM-5
& 358B
& 75.9
& 72.7
& \multicolumn{1}{c}{--}
& \multicolumn{1}{c}{--}
& \multicolumn{1}{c}{--} \\

DeepSeek-V4-Pro-Max
& 1.6T
& {\bfseries 83.4}
& \multicolumn{1}{c}{--}
& \multicolumn{1}{c}{--}
& {\bfseries 80.0}
& \multicolumn{1}{c}{--} \\

GPT-5 High
& --
& 54.9
& 63.0
& 77.9
& 75.0
& 76.4 \\

\midrule
\multicolumn{7}{c}{
    \emph{\textbf{Search Agent ($\sim$30B)}}
} \\
\midrule

MiroThinker-1.7-mini
& 30B
& {\bfseries 67.9}
& {\bfseries 72.3}
& \multicolumn{1}{c}{--}
& 57.2
& {\bfseries 80.3} \\

RedSearcher
& 30B
& 57.4
& 58.2
& \multicolumn{1}{c}{--}
& \multicolumn{1}{c}{--}
& 80.1 \\

DeepMiner
& 32B
& 33.5
& 40.1
& 62.0
& \multicolumn{1}{c}{--}
& 58.7 \\

Tongyi-DeepResearch
& 30B
& 43.4\rlap{\textsuperscript{*}}
& 46.7\rlap{\textsuperscript{*}}
& {\bfseries 75.0}
& \multicolumn{1}{c}{--}
& 70.9 \\

OpenSeeker
& 30B
& 29.5\rlap{\textsuperscript{*}}
& 48.4\rlap{\textsuperscript{*}}
& 74.0
& \multicolumn{1}{c}{--}
& \multicolumn{1}{c}{--} \\

\midrule
\multicolumn{7}{c}{
    \emph{\textbf{Search Agent (4B)}}
} \\
\midrule

QUEST-4B
& 4B
& 40.0
& \multicolumn{1}{c}{--}
& \multicolumn{1}{c}{--}
& \multicolumn{1}{c}{--}
& 77.7 \\

DR-Venus
& 4B
& 29.1\rlap{\textsuperscript{*}}
& 37.7\rlap{\textsuperscript{*}}
& 74.7
& 40.7
& 64.4 \\

AgentCPM-Explore
& 4B
& 24.1\rlap{\textsuperscript{*}}
& 29.1\rlap{\textsuperscript{*}}
& 70.0
& 34.0
& 63.9 \\

\midrule

\textbf{ABSeeker}
& 4B
& {37.3\rlap{\textsuperscript{*}} / \bfseries 55.3}
& {39.1\rlap{\textsuperscript{*}} / \bfseries 52.9}
& {\bfseries 77.0}
& {\bfseries 46.0}
& {\bfseries 81.6} \\

\bottomrule
\end{tabular}}
\end{table}

\noindent \textbf{Main Results.}
Table~\ref{tab:main-results} compares ABSeeker with foundation models and representative search agents across five benchmarks. Among 4B search agents, ABSeeker achieves the best performance on every benchmark, scoring \textbf{55.3\%} on BrowseComp, \textbf{52.9\%} on BrowseComp-ZH, \textbf{77.0\%} on xbench-2505, \textbf{46.0\%} on xbench-2510, and \textbf{81.6\%} on GAIA-text. Despite its smaller model size, ABSeeker also remains competitive with substantially larger search agents. It outperforms all reported 30B agents on xbench-2505 and GAIA-text, while surpassing several 30B systems on both BrowseComp and BrowseComp-ZH. Notably, although our method is trained exclusively on BrowseComp-style questions, it generalizes effectively to xbench and GAIA, indicating strong cross-benchmark generalization. These results demonstrate that through answer-backtracked step-level credit assignment---which rewards useful actions even in failed trajectories and penalizes redundant or erroneous behaviors in successful ones---the model learns to search more efficiently and more deliberately, validating the effectiveness of our approach.

\begin{figure*}[t]
\centering
\includegraphics[width=0.9\textwidth]{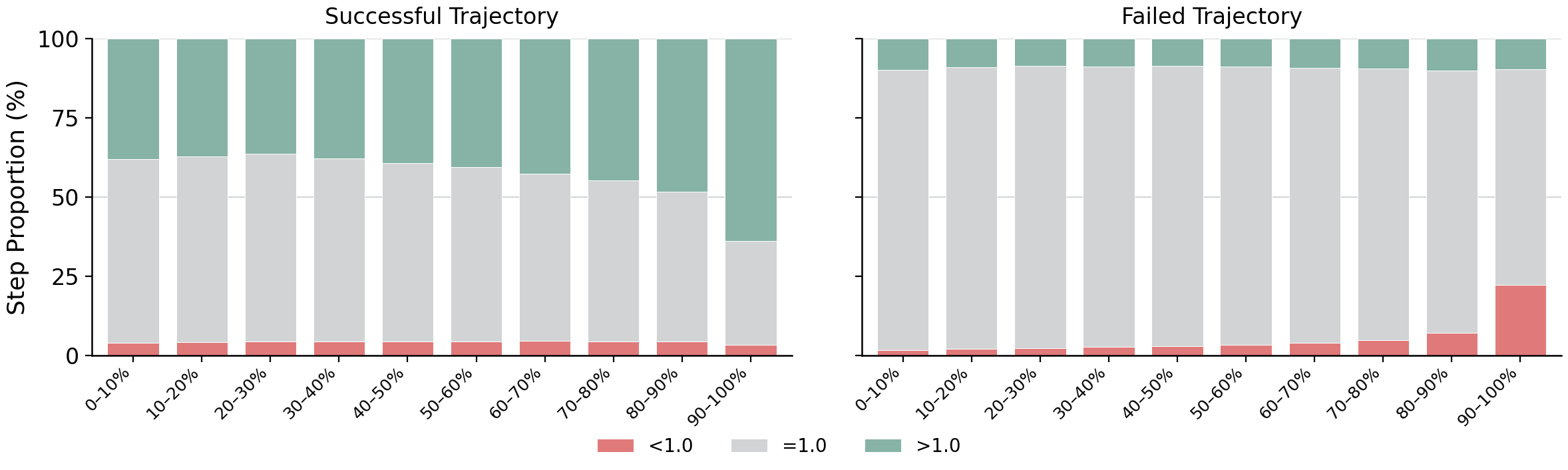}
\caption{Distribution of step rewards across the 8.5K SFT trajectories. The left half shows successful trajectories, while the right half shows failed trajectories. Each bar represents one-tenth of the steps in a trajectory, ordered from earliest to latest; for example, 0--10\% represents the first 10\% of steps in a trajectory. Red, gray, and green denote low-quality steps with $r_t<1.0$, neutral steps with $r_t=1.0$, and high-quality steps with $r_t>1.0$, respectively.}
\label{fig:reward-distribution}
\end{figure*}

\noindent \textbf{Reward Distribution Analysis.}
We further analyze the step-reward distribution over the 8.5K trajectories used for \textbf{ABC-SFT}. As shown in Figure~\ref{fig:reward-distribution}, even successful trajectories contain approximately 4\% low-quality steps with rewards below $1.0$. More importantly, nearly 10\% of the steps in failed trajectories receive rewards above $1.0$, indicating that they still discover or verify useful clues despite ultimately producing an incorrect answer. Trajectory-level supervision would assign the same outcome signal to all of these steps, thereby reinforcing erroneous actions in successful trajectories while penalizing useful actions in failed ones. In contrast, \textbf{Answer-Backtracked Credit Assignment} evaluates each step independently, preserving productive exploration and suppressing incorrect decisions. This precise supervision accounts for the consistent gains over the corresponding baselines. See Section~\ref{sec:ablation} for a detailed ablation analysis.

\begin{figure}[t]
\centering

\begin{minipage}[t]{0.65\textwidth}
  \centering
  \includegraphics[width=0.49\linewidth]{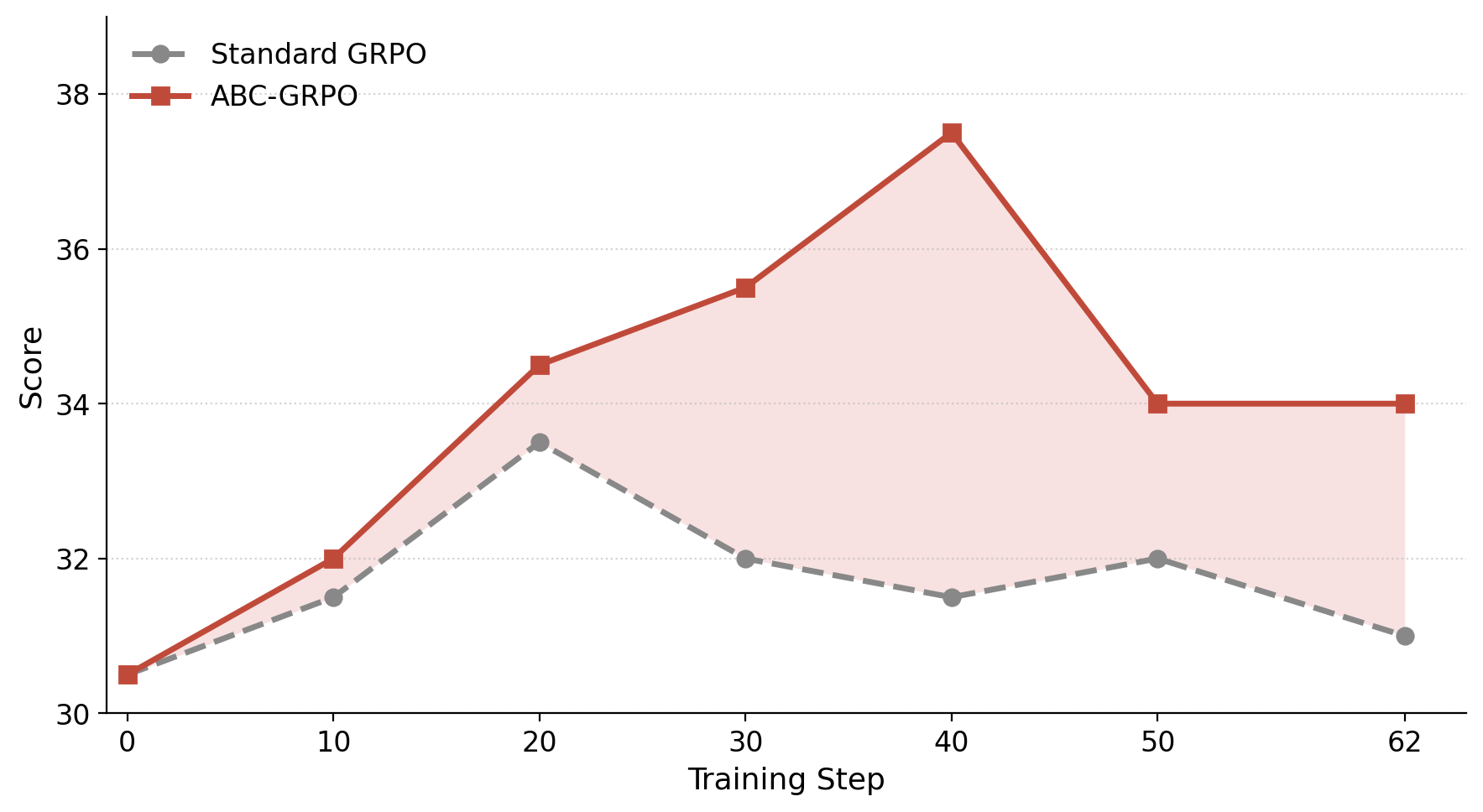}
  \hfill
  \includegraphics[width=0.49\linewidth]{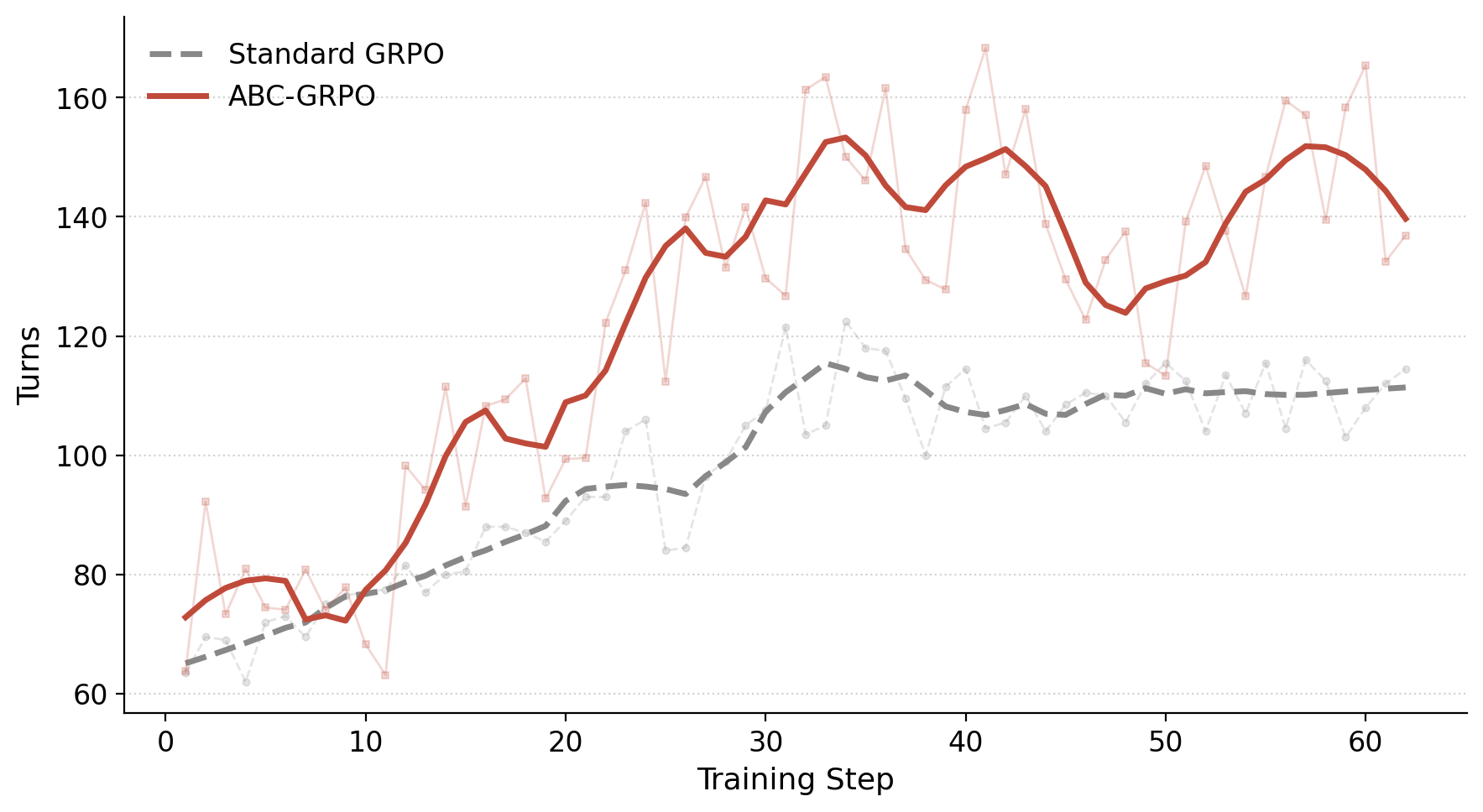}

  \captionsetup{
    justification=raggedright,
    singlelinecheck=false,
    margin=0pt
  }
  \captionof{figure}{
    RL training dynamics.
    \textbf{Left}: Performance during training.
    \textbf{Right}: Average number of interaction turns.
  }
  \label{fig:rl-ablation}
\end{minipage}
\hfill
\begin{minipage}[t]{0.32\textwidth}
  \centering
  \includegraphics[width=\linewidth]{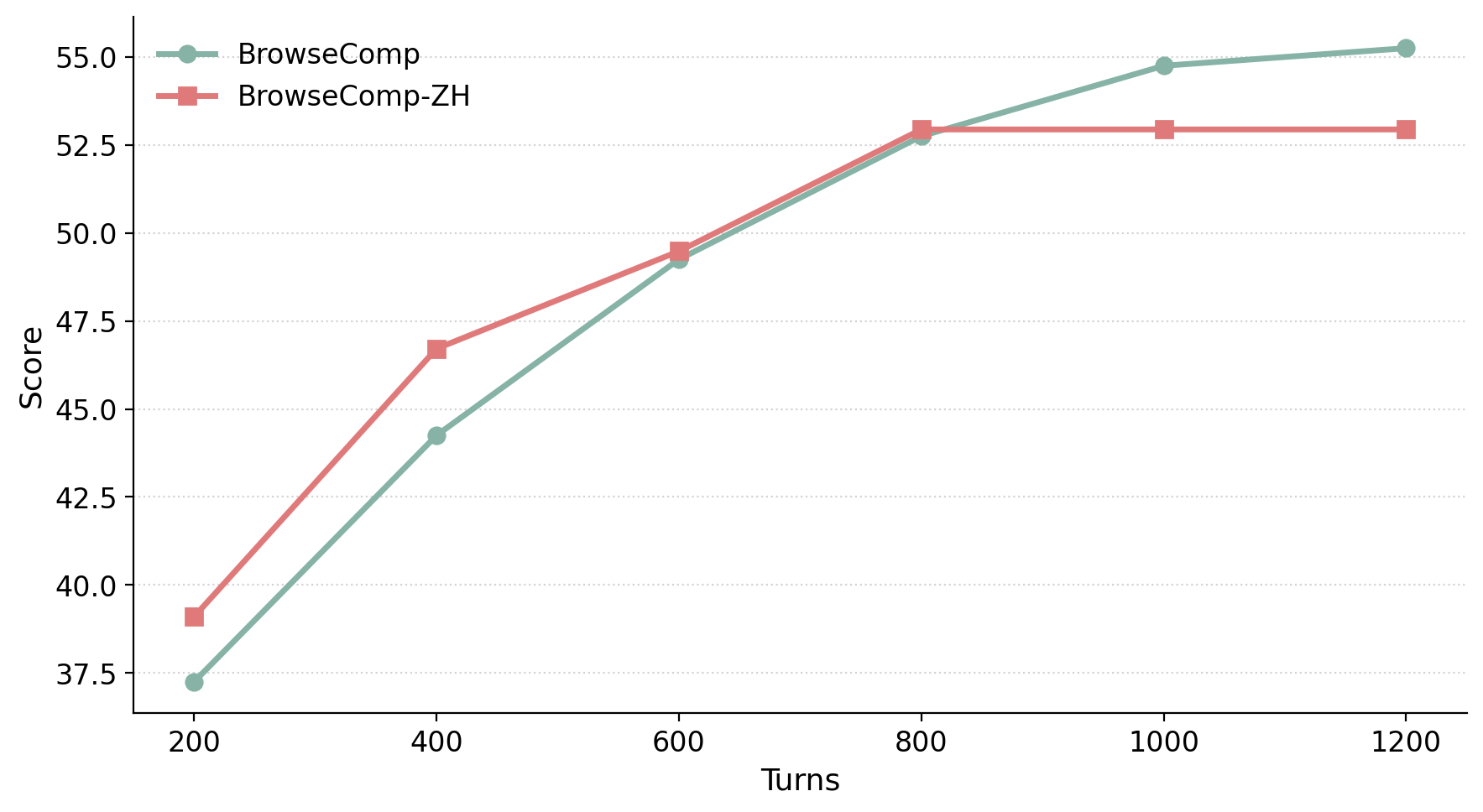}

  \captionsetup{
    justification=raggedright,
    singlelinecheck=false,
    margin=0pt
  }
  \captionof{figure}{
    Performance under different context budgets.
  }
  \label{fig:context-management}
\end{minipage}

\end{figure}

\noindent \textbf{RL Training Dynamics.}
To better understand the source of these gains, we compare \textbf{ABC-GRPO} with standard trajectory-level GRPO during training, both evaluated on a validation set of 200 randomly sampled BrowseComp questions. As shown in Figure~\ref{fig:rl-ablation}, \textbf{ABC-GRPO} achieves consistently stronger BrowseComp performance after training begins while producing longer search trajectories. Together, these results show that step-level credit assignment improves both search accuracy and exploratory behavior.

\noindent \textbf{Effect of Context Management.}
Following MiroThinker~\citep{team2026mirothinker} and LongSeeker~\citep{lu2026longseeker}, we set the maximum context length to 256K tokens and apply the discard-all strategy for up to five rounds on BrowseComp and BrowseComp-ZH. As shown in Figure~\ref{fig:context-management}, ABSeeker improves from 37.3\% to 55.3\% on BrowseComp and from 39.1\% to 52.9\% on BrowseComp-ZH.
% The earlier saturation on BrowseComp-ZH may reflect differences in benchmark quality: BrowseComp questions tend to have more discriminative constraints and uniquely supported answers, making the agent less likely to converge on plausible but incorrect alternatives as the search continues.

\subsection{Ablation Studies}
\label{sec:ablation}

\begin{table}[t]
\centering
\caption{Ablation study of ABC-SFT and ABC-GRPO. Results are evaluated without context management.}
\label{tab:ablation}
\resizebox{\linewidth}{!}{%
\begin{tabular}{@{}lccccc@{}}
\toprule
\textbf{Model} & \textbf{BrowseComp} & \textbf{BrowseComp-ZH} & \textbf{xbench-2505} & \textbf{xbench-2510} & \textbf{GAIA-text} \\
\midrule
\multicolumn{6}{l}{Qwen3.5-4B} \\
\midrule
+Standard SFT
& 28.5 & 30.4 & 73.0 & 27.0 & 66.0 \\

+ABC-SFT
& 30.8 & 31.8 & 72.0 & 35.0 & 72.8 \\

\midrule
\multicolumn{6}{l}{ABSeeker-4B-SFT} \\
\midrule
+Standard GRPO
& 33.5 & 36.3 & 75.0 & 41.0 & 77.7 \\

+ABC-GRPO
& \textbf{37.3} & \textbf{39.1} & \textbf{77.0} & \textbf{46.0} & \textbf{81.6} \\
\bottomrule
\end{tabular}}
\end{table}

Table~\ref{tab:ablation} evaluates ABC-SFT and ABC-GRPO across all five benchmarks, with all methods tested without context management. Compared with standard SFT, ABC-SFT improves performance on BrowseComp, BrowseComp-ZH, xbench-2510, and GAIA-text, while remaining comparable on xbench-2505. Building on this initialization, ABC-GRPO consistently outperforms standard trajectory-level GRPO across all benchmarks. These results demonstrate that fine-grained step-level credit assignment improves both SFT and RL by enabling the model to emphasize useful actions and suppress erroneous ones throughout training.

% ---------------------------------------------------------------
\section{Conclusion}
% ---------------------------------------------------------------
\label{sec:conclusion}

We present \textbf{Answer-Backtracked Credit Assignment (ABC)}, a fine-grained credit assignment framework for training long-horizon search agents. Instead of treating all steps within a trajectory uniformly, ABC recovers intermediate evidence clues from verified answers and uses them to evaluate each search step. Based on the resulting clue-anchored rewards, we develop \textit{ABC-SFT}, which reweights the loss of each turn, and \textit{ABC-GRPO}, which uses step-level scores as rewards in GRPO. These methods reward useful actions in failed trajectories while suppressing erroneous or redundant actions in successful ones. Building on this framework, we train \textbf{ABSeeker} based on Qwen3.5-4B. Experiments across BrowseComp, BrowseComp-ZH, xbench, and GAIA-text show that ABSeeker outperforms same-scale baselines and remains competitive with substantially larger search agents. Further analyses demonstrate that answer-backtracked step-level supervision improves reward quality, training dynamics, and long-horizon exploration, highlighting the importance of explicit process supervision for scalable search-agent training.

\textbf{Future Work.}
Due to computational constraints, our experiments focus on a compact 4B model. A natural next step is to scale ABSeeker to larger backbone models and examine whether answer-backtracked credit assignment brings stronger gains under higher model capacity. Beyond web search, we also plan to extend this framework to other long-horizon agent tasks where final outcomes can be backtracked into intermediate evidence, subgoals, or decision points to provide fine-grained process supervision.

\medskip
{
\bibliographystyle{plainnat}
\bibliography{ref}
}

\clearpage
\appendix

\section{Training Details}
\label{app:training}

\textbf{Supervised Fine-Tuning.}
We implement SFT using Slime and initialize the model from Qwen3.5-4B. We randomly select 8.5K trajectories from OpenSeeker~\citep{du2026openseeker}, consisting of 5.5K correct and 3.0K incorrect trajectories. We train on these 8.5K scored trajectories for 3 epochs with a global batch size of 64. The optimizer is Adam with a learning rate of $5\times10^{-5}$, cosine decay, a warmup ratio of $0.1$, a minimum learning rate of $1\times10^{-6}$, and weight decay of $0.1$. We use tensor model parallel size 2 and context parallel size 8 for long-context training. Tool responses are masked from the loss. For ABC-SFT, we map each step reward to a loss weight by $w(r_t)=2\sigma(2(r_t-1))$, where the neutral reward $r_t=1.0$ is mapped to weight $1.0$. 

\textbf{Reinforcement Learning.}
We implement RL using veRL and initialize from the SFT checkpoint. We train on 1,000 questions filtered by the number of interaction turns, of which 200 have fewer than 100 turns and the remaining 800 have at least 100 turns. We sample 8 rollouts per question with a training batch size of 16. Each rollout allows up to 200 interaction turns. We use asynchronous rollout with 16 agent-loop workers. The rollout policy uses temperature $1.0$, top-$p=0.95$. We optimize with ABC-GRPO using clue-anchored rewards plus a format penalty. Rewards are normalized within each rollout group, and discounted step-level advantages are computed with $\gamma=0.25$. The actor learning rate is $1\times10^{-6}$ and the KL loss coefficient is $0.001$.

\section{Evaluation Details}
\label{app:evaluation}
Due to computational constraints, for the RL validation set, we sample 200 BrowseComp questions from the full set using random seed 42. To reduce evaluation variance, we run each evaluation three times and report the averaged score. All evaluations allow up to 200 interaction turns and use DeepSeek-V4-Flash~\citep{xu2026deepseek} as the evaluation model with the prompt in Appendix~\ref{app:prompt-evaluation} to judge final-answer correctness.

\section{Prompt Templates}
\label{app:prompts}
\subsection{Answer-Backtracked Clue Recovery}
\label{app:prompt-recovery}

This prompt recovers intermediate clues from the query and verified answer.

\begin{promptbox}{Prompt for Answer-Backtracked Clue Recovery}

\textbf{System:}

You are an expert investigator. Use search\_web and visit\_web to gather evidence.
Do NOT provide the final answer directly. Extract intermediate clues, such as dates, places, and events, from the question, search for them first, then piece together to verify the answer.

\vspace{6pt}
\textbf{User:}

Reconstruct cite-backed steps that prove the answer is correct.

\vspace{4pt}
Question: \{query\}

Known Answer: \{answer\}

\vspace{4pt}
Your task: Find the intermediate clues that lead to the correct answer. These clues are usually conditions in the question, such as events, dates, places, and entities. Search for them first, then verify the answer. You MUST explicitly mention these intermediate clues in your final answer and explain how they support the answer.

\end{promptbox}

\subsection{Clue-Anchored Step Scoring}
\label{app:prompt-scoring}

This prompt scores each trajectory step using the recovered clue set.

\begin{promptbox}{Prompt for Clue-Anchored Step Scoring}

\textbf{System:}

You are a trajectory scorer. Your task is to evaluate each step in an AI agent's problem-solving trajectory based on the scoring rubric.

Focus on INTERMEDIATE STEPS and KEY ENTITIES. Base your scoring ONLY on the step's content, including Context, Reasoning, and Tool Call.

\vspace{6pt}
\textbf{User:}

\textbf{Scoring Rubric:}

Criteria 2--4 apply only to clues listed in ``Core Clues'' below. Each step starts from base score 1.0. Multiple criteria may apply in one step.

\vspace{4pt}
1. \textbf{Baseline (1.0):} Reasonable progress without an obvious error gets the base score 1.0.

2. \textbf{Finds or verifies a correct clue (+0.8):} The step discovers, investigates, or verifies a clue that matches the core clue set.

3. \textbf{Correctly rules out a wrong candidate (+0.4):} The step correctly excludes, dismisses, or identifies as wrong an irrelevant or incorrect candidate.

4. \textbf{Incorrectly dismisses a correct clue (-0.8):} The step wrongly excludes or dismisses a clue that is correct according to the core clues. Exploring wrong candidates does NOT count as this.

5. \textbf{Correct final answer (+1.0):} Only when the step explicitly gives the final answer and it matches the Correct Answer below.

6. \textbf{Wrong final answer (-1.0):} Only when the step explicitly submits an incorrect final answer. If this applies, do NOT apply criterion 5.

\vspace{6pt}
\textbf{Question:}

\{query\}

\vspace{4pt}
\textbf{Correct Answer:}

\{correct\_answer\}

\vspace{4pt}
\textbf{Core Clues:}

\{core\_clues\_text\}

\vspace{4pt}
\textbf{Step to Score:}

Step Number: \{step\_num\}

Context: \{user\_content\}

Reasoning: \{reasoning\}

Tool Call: \{action\}

Tool Response: \{response\}

\vspace{6pt}
\textbf{Task:}

Score this step based on the rubric above.

Return ONLY a JSON object with this exact structure:

\{
``score'': $<$float between 0.0 and 2.0; start from 1.0, add/subtract applied deltas, then clip to [0, 2.0]$>$,

``criteria'': $<$array of int 1--6, the rubric criterion numbers that apply, e.g., [1] or [1, 2, 5]$>$,

``explanation'': $<$brief explanation. When applying criteria, MUST name the specific clue(s)/entity, e.g., clue X or entity Y$>$
\}

Important: Return ONLY valid JSON, no other text.

\end{promptbox}

\subsection{Evaluation Prompt}
\label{app:prompt-evaluation}

This prompt checks whether the predicted answer matches the ground truth.

\begin{promptbox}{Prompt for Final Answer Evaluation}

\textbf{System:}

You are an evaluator. Your ONLY task is to decide: does the Predicted Answer refer to the SAME entity or value as the Ground Truth?

\vspace{6pt}
\textbf{User:}

Question: \{query\}

Ground Truth: \{gt\}

Predicted Answer: \{pred\}

\vspace{4pt}
Output only a JSON object with this format:

\{``correct'': true/false, ``reason'': ``brief explanation''\}

\end{promptbox}

\end{document}